# SRCA - The Scalable Robotic Cloud Agents Architecture

Vasilis N. Remmas,
Konstantinos L. Panayiotou,
Emmanouil G. Tsardoulias,
Andreas L. Symeonidis
School of Electrical and Computer Engineering
Aristotle University of Thessaloniki - AUTH
Thessaloniki 5414, Greece

*Abstract*— **In an effort to penetrate the market at an affordable cost, consumer robots tend to provide limited processing capabilities, just enough to serve the purpose they have been designed for. However, a robot, in principle, should be able to interact and process the constantly increasing information streams generated from sensors or other devices. This would require the implementation of algorithms and mathematical models for the accurate processing of data volumes and significant computational resources. It is clear that as the data deluge continues to grow exponentially, deploying such algorithms on consumer robots will not be easy. Current work presents a cloud-based architecture that aims to offload computational resources from robots to a remote infrastructure, by utilizing and implementing cloud technologies. This way robots are allowed to enjoy functionality offered by complex algorithms that are executed on the cloud. The proposed system architecture allows developers and engineers not specialised in robotic implementation environments to utilize generic robotic algorithms and services off-the-shelf.**

## I. INTRODUCTION

Robotics is currently one of the most rapidly evolving domains: logistics applications, the automotive industry, factories, or even homes appliances incorporate robotics. Many researchers work towards equipping robots with intelligence, this way serving advanced functionality. However, intelligence in everyday applications is limited, due to the accompanying high costs of the required computational resources. Face, object, voice, and speech recognition, simultaneous localization and mapping (SLAM), navigation, path planning, and kinematic solvers are some of the typical cases where intelligence is needed.

Apart from the algorithm efficiency (error metrics), execution time and power consumption are also taken into account in order to decide if they can be used in a real system or not. Several algorithms which are executed in-robot require high power consumption that consumer robots not have, due to hardware limitations of their embedded computational resources, when compared to conventional desktop computers. Especially in cases where the robot is autonomous (relies on batteries as energy source), power consumption plays a critical role as it affects the maximum time of operation. Furthermore, as the amount of acquired data increases, both computational and power requirements of the algorithms increase. For example, current cameras provide higher resolutions than the past, resulting in larger image sizes and therefore in heavier image processing.

Based on the aforementioned concerns, infrastructures are being designed to virtually increase the available resources, even in cases of low cost robots. Nowadays, researchers use the cloud to provide Software-as-a-Service (SaaS) and Platform-as-a-Service (PaaS) solutions, allowing users to avoid large investment in hardware resources. This constitutes the largest advantage of cloud infrastructures, as they can provide huge amounts of processing power at low costs. As far as robots are concerned, they could easily offload and execute on the cloud algorithms that are not time critical, fact that would extend the robot operation time, as well as its overall AI capabilities.

The aim of the proposed SRCA architecture is to introduce a framework where researchers and engineers can submit, build and deploy robotic-oriented services on the cloud. This way, non-time-critical algorithms that are now executed in-robot can be offloaded in a cloud environment lifting computational power barriers, and enabling the extensibility of robotic functionality, regardless of the actual embedded platform(s) at hand. The SRCA architecture is the extension of the Cloud Agents architecture, presented in [1].

The paper is structured as follows. Chapter 2 contains information on existing architectures which aspire to solve the problem of offloading computational intensive algorithms from robot platforms to cloud infrastructures. In chapter 3 the SRCA architecture is presented, along with adopted technologies and tools. Submission, build, deployment and utilization procedures are also described in this chapter. Thereafter, chapter 4 presents the results of several experiments performed towards measuring and evaluating the performance and scalability of the system. Finally, in chapter 5 future work is described, oriented towards improved performance and Quality of Service.

## II. STATE OF THE ART

Robot manufacturers and distributors focus on reducing the cost of robots, resulting in limited capabilities. This leads to the development of new product versions in order to integrate new functionality, which otherwise could be simply updated, provided the correct software. The scientific community envisioned means to simplify the ways to update or enhance robots features. Platforms have been created through which a specific robot can download a program or an application and seamlessly execute it, two examples of which are RAPP [2][3] and RoboEarth [4], promoting the cloud-based robotics concept. Both RAPP and RoboEarth use the ROS [5] meta-operating system for implementing robotic controllers, in order to support the execution and offloading of robotic computations to the cloud, without the need of heavy modifications in already existing controller implementations.

The RAPP Framework was created within the context of RAPP Project (FP7-ICT-610947) and consists of two main tiers. The first concerns the software that should exist in the robot, so that it is considered RAPP-ready. The second part concerns the software executed on the cloud, denoted as RAPP Platform [3]. The services provided by the RAPP Platform are robotics-oriented and are utilized by RAPP applications by invoking API calls. Such services may include facial and object recognition, speech synthesis and speech recognition tasks, among others. Each RAPP developer can also create such a service and deploy it on the cloud. RAPP Platform provides services for the submission and deployment of Cloud Agents as well [1]. There, a Docker container for each service [6] is created, all the necessary dependencies are installed according to the input parameters provided by the developer and finally the container is deployed. Ultimately, and upon successful submission, Cloud Agent services are being forwarded out of the container and exposed via a web server, over HTTP1.1 protocol. The RAPP Cloud Agents architecture has several particularities. First of all, services are provided via HTTP POST requests at a web server that uses TCP-Sockets or ROS-websockets to communicate with the container that owns that specific Cloud Agent package. It is worth noting that there is one web server to serve all services. Also, the robot has to send a request to the platform in order to create and deploy the container which offers a new service.

Another interesting cloud robotics platform is Rapyuta, the core implementation of the RoboEarth project [7]. It contains three communication layers, a set of tasks, as well as a set of commands by which the user can manage the system. Services can be used not only by robotic devices, since the service calls are based on standard HTTP requests. This makes it even easier for developers to use it, as calling services is analogous to invoking simple functions. Finally, services can be utilized via C++ or Python API client libraries, provided by the framework.

An alternative approach that has been developed to solve the problem of offloading computational resources to the cloud is Cloudroid [8]. It uses Docker containers and Docker Swarm for their orchestration. In addition, it uses ROSBridge web sockets for exposing ROS resources to the world-wide web, in order for robots to invoke services by using the auto-generated stubs Cloudroid provides. A remarkable feature of this system is the implemented Quality of Service mechanisms, which improve the speed and the reliability of the overall framework. Nevertheless, even though SCRA and Cloudroid are similar, SCRA is built on Kubernetes instead of Docker Swarm, supports ROS and non-ROS code, and provides a range of client libraries in terms of programming languages instead of just WebSockets.

The aforementioned architectures differ from the proposed system architecture in several parts. First of all, the SRCA system utilizes one web server for each container and a reverse proxy server, in order to allow communication between the service and the robot. This way, each package that exposes one or more services is isolated. Furthermore, the robot or the robotic developer is not involved in the process of creating the appropriate environment that a service may require, but just calls a specific service using the generated API clients. The user is responsible to fully define the service at the provided SRCA UI. Finally, it uses Kubernetes [9] as container orchestration framework which is similar to Docker Swarm. A more detailed explanation of the architecture is described in the next section.

### III. SRCA ARCHITECTURE

The system consists of three distinct interconnected layers (Figure 1). The first concerns Kubernetes, responsible for the management and orchestration of the Docker containers. Kubernetes uses one node of the cluster as a Master Node, uptaking the task of directly communicating with the other nodes towards task allocation and assignment. Each node can host different robotic services assigned by the Master Node. The computational resources each service is allowed to use are also set by the Master Node. This implies modularity and scalability of the system and at the same time allows for full-range monitoring (nodes, pods, clusters, physical resources etc.).

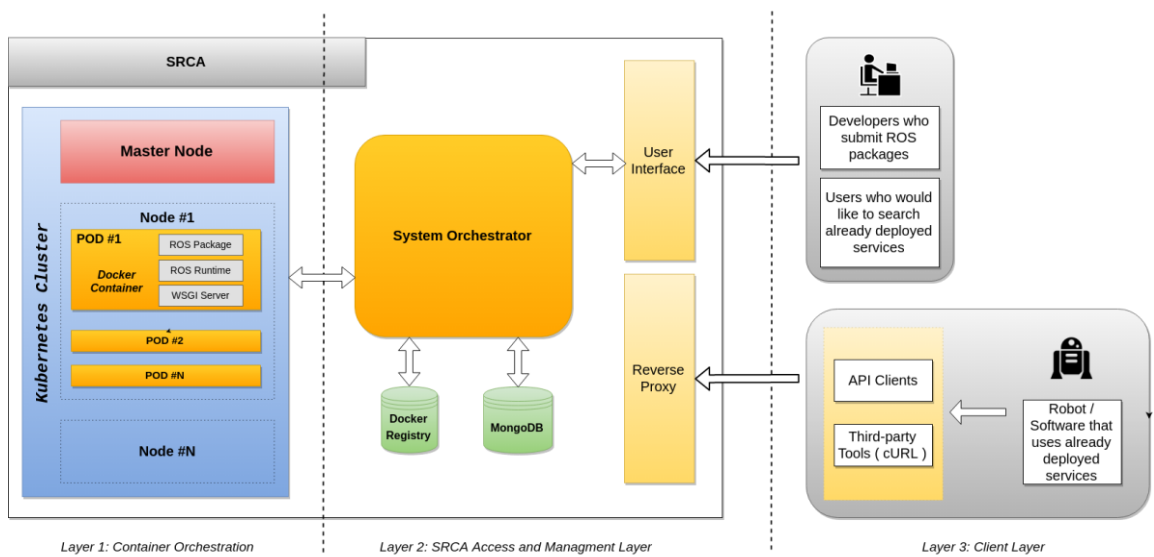

Figure 1: Architecture of the SRCA system

The second layer concerns the automated build and deployment procedures of ROS packages. This is the most important part of the system since it is responsible for the orchestration of all subsystems and also for the task allocation and assignment. It is implemented in NodeJS and integrates a web server (HTTP protocol), a reverse proxy server and a NoSQL database (MongoDB [10]), which are installed in the Master Node mentioned before. The web server utilizes Docker, Kubernetes and MongoDB APIs. The reversed proxy server is used in order to publicly expose the deployed services, making them accessible by the robots or software. Also, a web-based UI is implemented in this layer, enabling interactions between users and the cloud infrastructure. At this layer, the programming environment in which the ROS package needs to get built and executed is automatically created. Also, the uploaded by the user package is parsed, in order to confirm that it fulfils the necessary specifications. Furthermore, an automatically generated communication medium (an API client library) is created for each ROS package using Swagger tools, thus conforming to OpenAPI Specifications (OAS). Swagger gives the opportunity to generate the necessary files that an API client consists of. OAS defines a standard, programming language-agnostic interface description for REST APIs. This client library can be embedded into either a robot controller or an application and utilize the deployed cloud services.

The last layer of the system is the client layer. It involves the users, the robots and the software that utilize the system's features. Through the UI, users can upload ROS packages, adjust the computing resources to be used for each one (scaling), track the status of both building and deployment procedures, as well as have access to runtime logs. Also, one can search and find deployed ROS packages, retrieving necessary information about how to call the services each one contains. Finally, robots or software can use the generated API client libraries that the system provides or by employ third-party tools like cURL, in order to call already deployed services.

The main concept is that a developer can upload a ROS package accompanied by a configuration file. This file contains information about the automatic build and deployment of the ROS package in an isolated containerized environment existing on the cluster. It is worth mentioning that the system does not support pre-compiled Docker containers, since auto-generated files that should be included in the same container are involved. Upon successful deployment, proper URLs are automatically generated, via which any authorized user can make HTTP calls in order to use a specific service of this package. Also, an OpenAPI specifications file is generated along with API client libraries towards invocation from a range of programming languages like C++, Python and JavaScript. The programmer can manage his/her deployed services through the SRCA UI in order to scale it up or down (assign or remove deployed containers), review the logs relevant to the build and/or deployment procedures and retrieve further information about the deployment (e.g. the version of the package).

A more detailed explanation about how the system operates follows. Initially, the user uploads a ROS package along with a configuration file, written in YAML format. Figure 2 presents an example of a YAML file used to automatically build and deploy a ROS package named *test,* comprising two different functions defined in two different files.

This YAML file contains critical parameters, used to automatically build and deploy the ROS package. The definitions and roles of the main ones follow:

- **file:** a source code file containing one or more functions. The user has to create and provide these files and place them in a folder named *functions* in the ROS package directory. The files can only be developed in Python, as it's currently the only supported language. Their structure can be really simple like the script presented at Figure 3.

- **function:** a function the user must create in order to invoke one or more ROS services and get the returned value through the web server.

- **package:** can contain *apt-get*, *pip* and *npm* commands, i.e. package managers used to download and install specific packages, followed by the name of the software the ROS package requires to get properly built and executed.

- **command:** the appropriate command that will launch ROS Core and the necessary ROS nodes for the package to serve the contained services. This is the one and only command that the container will run at execution time.

```yaml
name: test
version: v1
environment: ROS
files:
- file_name: client.py
  functions:
  - name: add_two_ints
    arguments:
      params:
        a: integer
        b: integer
    http-method: post
    returns: string
- file_name: testfiles.py
  functions:
  - name: sendmyfile
    arguments:
      files:
        fa:
      params:
        a: integer
    http-method: post
  returns: file
packages:
 apt-get: net-tools vim
 pip: numpy
 npm: underscore
command: roslaunch test launch.launch
```

Figure 2: Example of YAML file used to build and deploy package at SRCA

```python
#!/usr/bin/env python
import sys
import rospy
from test.srv import *
def add_two_ints(x, y):
    rospy.wait_for_service('add_two_ints_srv')
    try:
    add_two_ints_srv = rospy.ServiceProxy('add_two_ints_srv', AddTwoInts)
    resp1 = add_two_ints_srv(x, y)
    return resp1.sum
    except rospy.ServiceException, e:
    print "Service call failed: %s"%e
```

Figure 3: Structure of a Python script containing one function

Upon uploading the aforementioned files in a ZIP file, the developer waits until the functions are deployed. Meanwhile, the system performs the activities presented in Figure 5. Initially, it performs appropriate validation checks concerning the file's existence and the configuration file structure and schema. The next step is to generate a Docker file that will be used to build the ROS package, along with some configuration files needed to get everything operating correctly inside the container. Also, a Web Server Gateway Interface (WSGI) is generated in which the above functions are binded. The system uses a Gunicorn web server to host and run this WSGI application combined with three Eventlet workers, aiming at supporting more concurrent HTTP requests. This web server is running inside every single container and it is different for each package, thus a mechanism to bind internal to external ports will be needed (e.g. a reverse proxy). In order to achieve faster and parallel building of packages, the system supports building these images at different Kubernetes Pods.

After the successful completion of the build process, the system creates a **Kubernetes Deployment** which undertakes the deployment of this container at the Cluster, identified by username. In order to achieve that, **Kubernetes Namespaces** are used to craft a proper namespace for every different user of the system, making it easier to debug errors. Next, a **Kubernetes**

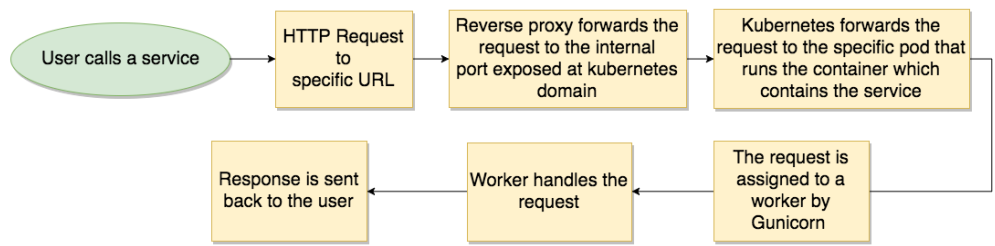

Figure 4: The path which an HTTP Request follows at SRCA

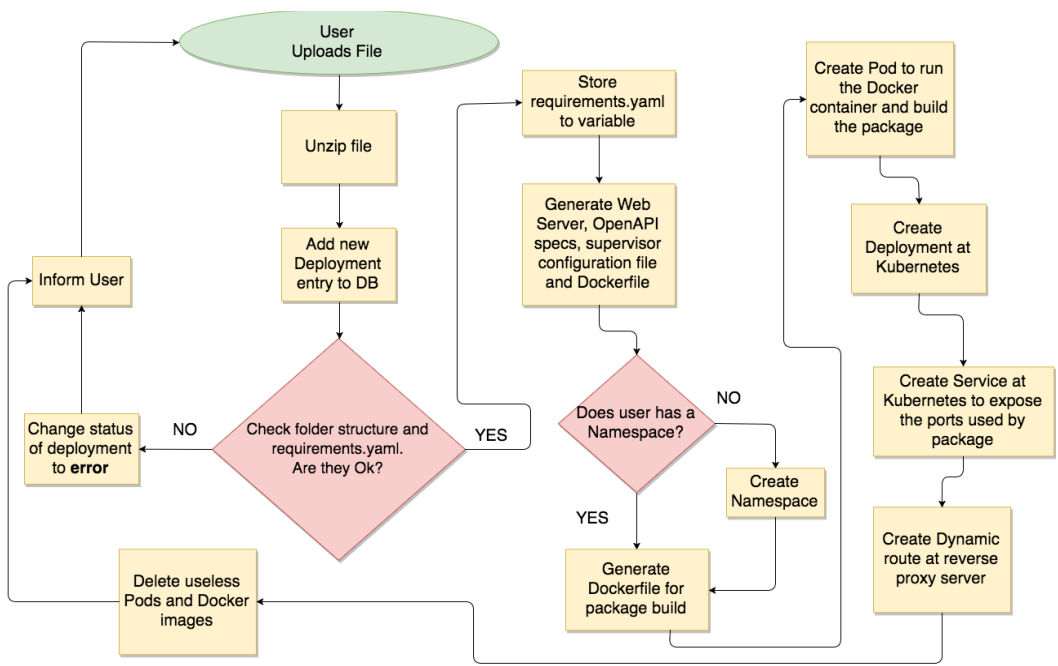

Figure 5: Flow chart of automatic build and deployment procedures

**Service** is created and is used to forward the port of the Gunicorn Web Server existing in the container. This port is available only to entities belonging in the same network as the Master Node. A reverse proxy server has been integrated, in order to dynamically create routes that connect an internal port like the one exposed by the Kubernetes Service with an external user-friendly URL. This reverse proxy server is again a custom implementation using the NodeJS web server framework. The route that an HTTP request follows in the SRCA is presented in Figure 4.

During this procedure, the database is getting updated in order to store the status of this process. This NoSQL database consists of four collections, responsible for storing I) information about the users, II) the deployments created by the users (including the services that the specific package contains), III) the devices that they own and IV) user groups with different permissions (e.g. number of pods that are allowed to use for a specific deployment).

Finally, the system cleans up any useless Docker images and Kubernetes Pods that have been created and informs the user that the functions are ready for utilization. The user can then modify this Deployment through the aforementioned SRCA UI.

## IV. EXPERIMENTS ON SYSTEM PERFORMANCE

The system was tested using a Cluster of three (3) Nodes. Each Node has 4GB RAM, 40GB HDD and 1 CPU core. Kubernetes and Docker were installed in each, in order to create a Kubernetes Cluster. Also, an instance of MongoDB was installed at the Master Node along with the NodeJS web server described in the previous chapter. The experiments include a ROS package that has only one function named ***add_two_ints()***. This function takes as input arguments two numerical values and returns the sum of them. We concluded in using this function at the experiments, due to the fact that it is not a computationally intensive operation. In this way, we have measured the performance and the response time of the system, instead of the computational power of the current cluster's resources. The reader should take into account that Kubernetes Nodes may be similar to a standard PC regarding the computational power, thus we expect that a multi-node ROS package will behave the same in both configurations. Different experiments we conducted are for 1-3 Pods and 1-3 eventlet workers for 500 simultaneously HTTP requests using cURL. Below, a list containing the metrics, as defined by cURL, is presented:

- **time_connect**: The time, in seconds, from the start until the TCP connect to the remote host (or proxy) was completed.

- **time_pretransfer**: The time, in seconds, from the start until the file transfer was just about to begin. This includes all pre-transfer commands and negotiations that are specific to the particular protocol(s) involved.

- **time_start_transfer**: The time, in seconds, from the start until the first byte was just about to be transferred. This includes time_pretransfer and also the time the server needed to calculate the result.

- **time_total**: The total time, in seconds, that the full operation lasted.

- **total_start_transfer:** Time difference between time_total and time_start_transfer.

The diagrams in Figure 6 depict the differences between 1-3 Pods and 3 worker processes. The results presented in Table I conclude that as the number of Pods increases, the mean time difference between *time_total* and *time_start_transfer* decreases. This means that the server can serve faster and smoother concurrent requests. Furthermore, the same experiments have been performed for a variable number of worker processes (1-3) and the results are presented in Table II. It is noticeable that there are combinations with less computer resources that perform better. e.g. using 2 Pods and 1 worker performs better than using 2 Pods and 2-3 workers. This may occur due to the fact that the specific cluster is not powerful enough to assign without delay the requests which arrive at the web server to the specific workers.

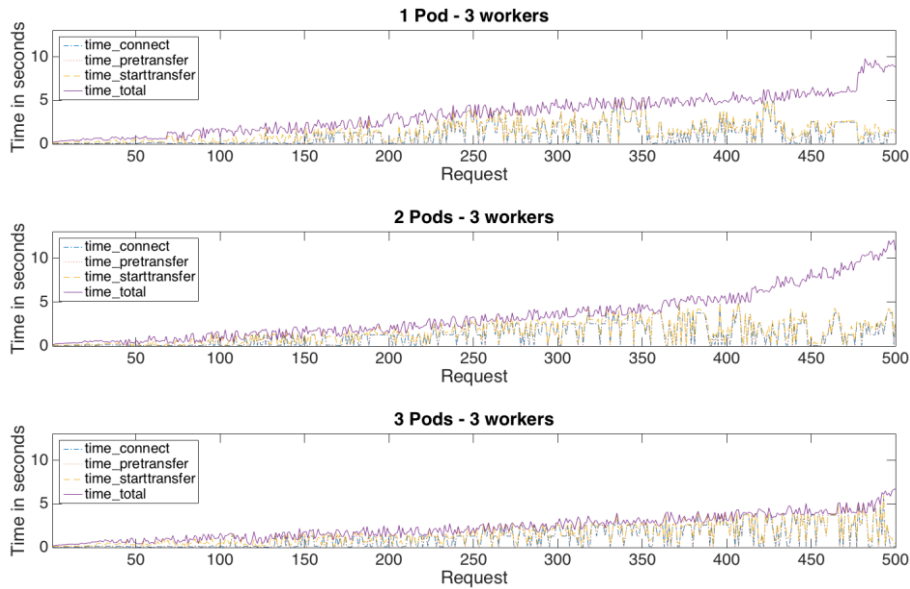

Figure 6: System performance for 500 concurrent requests

It is worth mentioning that the maximum number of worker processes for each web server was equal to three, as it has been experimentally noticed that it is optimized for the specific setup of the cluster. However, a feature could be added in the future which will enable the user to change that number.

TABLE I. TIMES FOR 3 WORKERS AND 500 REQUESTS

| # of Pods | Average Time | | | | |
|---|---|---|---|---|---|
| | connect | pre transfer | start_transfer | total | total - start_transfer |
| 1 | 1.03618 | 1.03636 | 1.43266 | 3.46107 | 2.02840 |
| 2 | 1.20848 | 1.20860 | 1.54866 | 3.50012 | 1.95146 |
| 3 | 1.17671 | 1.17688 | 1.51506 | 2.32103 | 0.80597 |

TABLE II. TIME COMPARISON BETWEEN DIFFERENT WORKERS FOR 500 REQUESTS

| # of Pods | Average Time Difference Between total and start_transfer | | |
|---|---|---|---|
| | 1 worker | 2 workers | 3 workers |
| 1 | 2.88665 | 2.65352 | 2.02840 |
| 2 | 1.72741 | 2.59609 | 1.95146 |
| 3 | 1.43738 | 1.58541 | 0.80597 |

## V. CONCLUSION / FUTURE WORK

The proposed SRCA architecture is based on modern and state-of-the-art technologies and tools, which evidently have large potentials to evolve and mitigate in common everyday procedures. SRCA is an architecture that promotes the Cloud Robotics concept, by allowing robotic experts to offer services for deployment by simple users or robots, in a structured and scalable way.

Currently, Kubernetes can handle 5,000 nodes and 150,000 pods, which makes SRCA extremely scalable. Also, using Kubernetes allows the proposed architecture to be deployed in physical cloud infrastructures like the one used for the experiments. In addition, using containers the way described above and decoupling the main components of the system achieve low granularity of the architecture. This makes the system perform smoother its operations and be more reliable. Furthermore, it should be clearly stated that the proposed system (as well as any other cloud-based service providing system) cannot be used for time critical operations. This means that a Pod cannot handle real time motion control or visual assisted manipulation, but functions the robot has the luxury to wait for their conclusion, since their response time cannot be known due to network delays. Several improvements can be performed as future work, in order to provide better Quality of Service and extended functionalities.

First of all, the system should be benchmarked with realistic (for robotics) services such as 2D/3D path planning, object detection/recognition, human skeleton detection and others. Furthermore, ROS uses streams over TCP sockets to allow communication between nodes (TCPROS). There are many cases where a user/robot needs to stream data to services (and not just call them once), thus maintaining an open channel of communication. This can be achieved through several protocols over TCP like websockets, MQTT, WAMP, or any other bidirectional network channel. Nevertheless, all the tools used to implement SRCA intra-layer communication interfaces support TCP sockets, thus the implementation of this feature does not require architecturally redesign the system.

Furthermore, upon code submission from a developer, the system creates a Docker image containing the entire package. This image is deployed on some Pods, but no communication means exists between them, thus each Pod executes the entire package separately. Nevertheless, ROS architecture allows the creation of more than one ROS nodes, which can communicate with each other. One can take advantage of Kubernetes' ability to create networks of Pods that can communicate with each other, resulting in the easier scaling of ROS Nodes with large workload, instead of scaling the entire package.

Finally, a registry is used to store the Docker images, which are currently being exploited only during their deployment process. The implementation of a subsystem to enable reusability of existing images would be a useful addition, especially for saving deployment time. This could considerably shorten the building time of a new image, in case it has similar specifications to another that already exists in the registry.